\documentclass[a4paper,10pt]{article}
\usepackage{ifthen}
\newboolean{review}
\setboolean{review}{false}

\usepackage[nobiblatex]{xurl}
\usepackage[breaklinks,colorlinks]{hyperref}
\usepackage{doi} 
\usepackage[hyperref=true,natbib,backend=bibtex,style=ext-authoryear-comp,maxnames=2,giveninits=true]{biblatex}
\DeclareFieldFormat{labelnumberwidth}{#1\adddot}
\DeclareNameAlias{sortname}{family-given}
\DeclareNameAlias{author}{family-given} 
\DefineBibliographyStrings{english}{and = {\&}} 
\DeclareFieldFormat
  [misc,article,inbook,incollection,inproceedings,unpublished]
  {titlecase:title}{\MakeSentenceCase*{#1}} 
\DeclareFieldFormat[misc,article,inbook,incollection,inproceedings,thesis,book,unpublished]{title}{#1} 
\renewbibmacro{in:}{%
  \ifboolexpr{%
     test {\ifentrytype{article}}%
     or
     test {\ifentrytype{inproceedings}}%
  }{}{\printtext{\bibstring{in}\intitlepunct}}%
} 
\AtEveryBibitem{\clearname{editor}} 
\DeclareFieldFormat[article]{volume}{#1\addcolon\space}
\DeclareFieldFormat[article]{pages}{#1} 
\AtEveryBibitem{%
  \ifentrytype{article}{%
    \clearfield{number}%
    \clearfield{issue}%
  }{}%
} 

\usepackage[dvipsnames]{xcolor}
\hypersetup{colorlinks=true,citecolor=blue,urlcolor=blue,linkcolor=blue,menucolor=blue}

\usepackage[capitalize]{cleveref}

\usepackage{afterpage} 

\usepackage{graphicx} 
\graphicspath{{Figs/}}

\usepackage{subfigure} 
\usepackage[bf]{caption} 
\usepackage{multirow}
\usepackage{rotating}

\usepackage{siunitx,array,booktabs,ragged2e}
\newrobustcmd\B{\DeclareFontSeriesDefault[rm]{bf}{b}\bfseries}
\newrobustcmd\F{\DeclareFontSeriesDefault[rm]{bf}{b}\bfseries\color[rgb]{0, 0, 0.6}}
\newrobustcmd\W{\DeclareFontSeriesDefault[rm]{bf}{b}\bfseries\color[rgb]{0.5, 0, 0}}

\usepackage[a4paper,portrait,margin=2cm]{geometry}

\usepackage{times} 
\usepackage{amsfonts}   
\usepackage{dsfont}
\usepackage{nicefrac}       
\usepackage{microtype}      

\usepackage[nodayofweek]{datetime} 

\usepackage{lineno}
\usepackage{enumitem} 
\setlist[itemize]{topsep=0pt,parsep=0pt,itemsep=0pt,leftmargin=1.8em}
\setlist[enumerate]{topsep=0pt,parsep=0pt,itemsep=0pt,leftmargin=2.4em}

\usepackage{multicol}

\usepackage{xspace}

\newcommand{\ie}{{\it i.e.},\xspace}

\newcommand{\hide}[1]{}

\newlength{\tabsecskip}
\newcommand{\figscale}{0.4}

\makeatletter
  \renewcommand\maketitle{\hfill \titlebibdata\\\par\noindent {\Large \bf \@title}\\[-1mm]\par \noindent {\large \bf \authorA}\hfill {\tt \emailA}\\[0.8mm] \noindent {\large \bf \authorB}\hfill {\tt \emailB}\\[0.8mm]\address}
\makeatother

\begin{document}

\title{HAND: A Biologically-Inspired Activation Function that Improves Generalisation and Sample Efficiency in Image Classification}
\newcommand{\authorA}{Michael W. Spratling}
\newcommand{\authorB}{Heiko H. Sch{\"u}tt}
\newcommand{\address}{University of Luxembourg, Department of Behavioural and Cognitive Sciences, L-4366 Esch-sur-Alzette, Luxembourg.}
\newcommand{\addressfull}{Department of Behavioural and Cognitive Sciences\\University of Luxembourg\\Maison Sciences Humaines\\11 Porte des Sciences\\L-4366 Esch-sur-Alzette\\Luxembourg}

\newcommand{\emailA}{{\tt michael.spratling@uni.lu}}
\newcommand{\emailB}{{\tt heiko.schutt@uni.lu}}

\newcommand{\orcidA}{0000-0001-9531-2813}
\newcommand{\orcidB}{0000-0002-2491-5710}
\newcommand{\titlebibdata}{\vspace*{-2ex}} 

\maketitle

\ifreview
\linenumbers
\fi

\begin{abstract}
  DNNs exhibit robustness and generalisation issues not seen in humans.
  They are also far less data-efficient learners, requiring considerably more training samples to accurately classify novel exemplars.
Inductive bias could help with these issues by providing in-built mechanisms to improve generalisation, and hence, reduce reliance on learning from data. 
We incorporate a biologically-inspired inductive bias into a new activation function, HAND (Homeostasis, Accelerating Nonlinearity, and Divisive-nomalisation), and show its effectiveness with CNNs trained on image classification. 
Using HAND a ConvNeXt-tiny required 25 training epochs to reach the same accuracy on ImageNet1k as the unmodified model achieved after 200 epochs. 
Consistent with the effects of an inductive bias, the performance gap reduced with training time and increased data augmentation.
When the volume of training data was reduced and unevenly distributed between classes (Long-tailed ImageNet) the improvements in accuracy were even larger and did not reduce with increased training time. 
Generalisation performance with the common-corruptions data, and the ability to reject samples from unknown classes, were uneffected or improved by HAND. 
Results generalised across CNN architectures and training data-sets.
HAND can, therefore, reduce the required training time and/or the required volume and variety of training data, helping to improve sample efficiency.
\end{abstract}

\paragraph{Code:} \url{https://codeberg.org/mwspratling/HAND}


\section{Introduction}
\label{sec-introduction}
\hide{
Early CNNs \citep{Fukushima88,RiesenhuberPoggio99b} were strongly inspired by neuroscience and, as a result, included multiple biologically-plausible mechanisms. For example, localised receptive fields (RFs), implemented using convolution, are a constraint designed to improve the learning of spatially-structured data, such as images, while reducing the number of parameters and mimicking the restricted RFs of biological neurons.
CNNs consist of repeating layers arranged in a hierarchy to resemble the cortical visual system. This hierarchical organisation allows 
neurons in the later stages to respond with increasing tolerance to location and scale to increasingly specific stimuli, as observed along the ventral stream \citep{GoodaleMilner92,KobatakeTanaka94,RustDiCarlo10}. Increased specificity and increased tolerance are produced, respectively, by the distinct operations performed by the integration and pooling layers \citep{RiesenhuberPoggio99b,Spratling05}, which were designed in early models to mimic the function of simple and complex cells found in primary visual cortex (V1) \citep{HubelWiesel62}.

While some of these characteristics have been inherited by more modern architectures \citep{Krizhevsky_etal12, ZagoruykoKomodakis16, Howard_etal19, He_etal16, Liu_etal22a, TanLe20}, work on engineering improved DNNs has tended to reduce biologically-inspired inductive bias. For example, 
pooling is often replaced by strided convolution removing the link to cortical complex cells, and the putative role of this type of neuron in increasing tolerance to appearance changes.
Similarly, transformers \citep{Dosovitskiy_etal20, Liu_etal22c, Jaegle_etal21, Liu_etal21} reduce the importance of the spatial arrangement of an image and localised RFs. 

}

A deep neural network (DNN) of sufficient size can approximate any arbitrary mapping from a set of inputs to a set of outputs \citep{Cybenko89, Hornik_etal89, Hornik91, KidgerLyons20, Zhang_etal17}. Hence, given sufficient training samples and time to search for the weights that produce the desired mapping, any image classification problem can be learnt without the need for strong, built-in, inductive biases. Inductive bias may, therefore, have limited influence on the final, optimised, solution learned from large data-sets. However, it can narrow down the range of putative solutions to guide learning towards a good solution in less time and with fewer training samples. Such behaviour would address multiple aims of frugal AI \citep{Girdhar_etal25}.

Despite these potential advantages, most work in deep learning has aimed to reduce inductive biases: relying on scaling-up (the size of the training data-set, the model, and the computational resources) to bring about improvements in performance. 
Similarly, most work in frugal AI is focused on enhancing the efficiency of models at inference time rather than the efficiency with which models are learnt (indeed most frugal AI methods increase training inefficiency).
In opposition to this overwhelming trend, some have advocated for the introduction of new biologically-inspired inductive biases into DNNs \citep{Zador19, Hassabis_etal17, Zhu_etal20, Malhotra_etal20, Linsley_etal25, Celeghin_etal23, Roy_etal24}. 
However, the only well-known examples of work in this direction modify the first stage of a DNN to simulate the function of the retina \citep{Stracke_etal25,Kim_etal15}, the lateral geniculate nucleus \citep{Bertoni_etal22}, or V1 \citep{Evans_etal22,Strisciuglio_etal20,Dapello_etal20}. The effects of such ``front-ends'' are  limited as they only change one stage out of many, and these effects could alternatively be produced by using an unmodified architecture with appropriate data-augmentations. More effective inductive biases are yet to be proposed. 

Here we propose a in-built inductive bias inspired by a fundamental constraint in vision: that only one feature can occupy each individual location in an image. 
This implies that there should be competition between neurons which represent different interpretations of the same image patch. Neurophysiological evidence suggests that this is the case in the cerebral cortex \citep{OlshausenField97,VinjeGallant00,CarandiniHeeger11,Lehky_etal05,Tolhurst_etal09}, and computational neuroscience suggests multiple different mechanisms by which this could be achieved \citep{SpratlingJohnson01b, Spratling14b, SpratlingJohnson04b, HarpurPrager94, Hoyer03, Zylberberg_etal11, Foldiak90, Rozell_etal08, Grossberg87, Song_etal00, Beyeler_etal19}. While a number of attempts to add competition to convolutional neural networks (CNNs) have been made in the past (\cref{sec-literature_review}), it remains missing from popular, contemporary, models.
We hypothesised that introducing competition into CNNs would enable better discrimination between distinct patterns, and hence, improve classification accuracy. Furthermore, we expected competition to reduce redundancy by encouraging neurons to represent distinct patterns. 
Current DNNs contain a high proportion of redundant neurons that serve no useful purpose \citep{Dalvi_etal20}. This fact is exploited in many methods of network pruning \citep{Wang_etal22,Nowak_etal23,Sehwag_etal20}, which aim to produce more efficient inference models by removing redundant neurons and/or connections from a pre-trained model. Such post-hoc pruning is inefficient as it relies on a large, redundant, model being trained first. Using competition to reduce redundancy should make more efficient use of the available neurons, without the need for pruning.

To test these hypotheses we designed a new activation function (HAND) that
uses divisive-normalisation: a simple competition mechanism that is a canonical computation in neurophysiology \citep{CarandiniHeeger11, CarandiniHeeger94, AlbrechtGeisler91, Heeger92, Heeger91, Wainwright_etal01, HeegerZemlianova20}. Our version of divisive-normalisation is applied across the channels to produce competition between neurons sharing the same RF location. HAND was used to replace the standard activation function in different CNN architectures. We evaluated the performance of HAND-based networks to their unmodified baselines using  three metrics: accuracy classifying the standard test dataset (``clean accuracy''), accuracy classifying the common corruptions data (``corrupt accuracy''), and ability to reject out-of-distribution samples (``OOD rejection''). We found that HAND typically led to improvements in all three performance metrics. This improvement was largest when the number of parameter updates was limited (\ie using few training epochs), or the training data was limited in variety (\ie using few training-time augmentations), or the training data was limited in volume (using reduced and imbalanced data): exactly those conditions in which a useful inductive bias should be most effective.
Divisive-normalisation therefore seems to act as an inductive bias that increases data efficiency while maintaining or improving generalisation and robustness. 

\section{Related Work}
\label{sec-literature_review}

Prior work has proposed many different mechanisms that modify a neuron's response based on the (previous or current) activity of other neurons. Only the existing literature that is most prominent or most relevant to the current work is discussed here.

\subsection{Normalisation Methods: Subtraction and Division}
\label{sec-review_standardisation}

Normalisation layers
have become common building-blocks in contemporary DNNs. These methods aim to standardise the mean and standard-deviation of neural activations in order to make training faster and more stable \citep{Santurkar_etal18}.
All these methods thus have the same formulation:
\begin{equation}
  z=\frac{
    y-\bar{y}}
  {\sqrt{\sigma + \frac{1}{n}\sum_{i\in \mathbb{P}}(y_i-\bar{y})^2}}  \;\;\;\;\;\;\;\;\;
  \bar{y}=\frac{1}{n}\sum_{i\in \mathbb{P}}y_i
  \label{eq-standardisation}
\end{equation}
Where $y$ and $z$ are scalars representing the activity of a single neuron before and after normalisation, respectively; $\sigma$ is a small non-negative constant used to prevent division-by-zero errors; and $\mathbb{P}$ is the set of $n$ values used to calculate the statistics. $\mathbb{P}$ differs between the different normalisation methods. Specifically, if a feature-map is represented by a tensor with dimensions (B,C,H,W), then
for BatchNorm \citep{IoffeSzegedy15} the statistics are calculated over (B,H,W) meaning that a separate mean and variance is calculated for each channel.
For LayerNorm \citep{Ba_etal16} the statistics are calculated over (C,H,W) meaning that a separate mean and variance is calculated for each sample in the batch.
For InstanceNorm \citep{Ulyanov_etal16} the statistics are calculated over (H,W).
GroupNorm \citep{WuHe18} is intermediate between LayerNorm and InstanceNorm meaning that a separate mean and variance is calculated for each predefined sub-set of channels of each sample in the batch.
An early version of InstanceNorm calculated the statistics in a local, 9x9, Gaussian-weighted spatial region \citep{Jarrett_etal09}, rather than across the whole height and width of the feature map.
These normalisation methods are typically followed by a learned affine transformation, which is omitted from the above equation.

\subsection{Divisive-normalisation Methods: Division Only}
\label{sec-review_div-norm}

Divisive-normalisation methods divide activity by a factor that is a weighted-sum of input values which have been raised to a power. Hence, a standard implementation is given by \citep{Heeger92,CarandiniHeeger11}:
\begin{equation}
  z=\frac{y}
  {\sigma + \bar{y}}  \;\;\;\;\;\;\;\;\;
  \bar{y}=\sum_{i\in \mathbb{P}}w_iy_i  \;\;\;\;\;\;\;\;\;
  y=x^\alpha
  \label{eq-div-norm}
\end{equation}
Where $x$, $y$ and $z$ are scalars representing the activity of a single neuron before, at an intermediate step, and after normalisation, respectively; $\sigma$ is a small non-negative semi-saturation constant; $\alpha$ is a non-negative parameter; and $\mathbb{P}$ defines the set of neighbouring neural activations that are used to calculate the weighted sum, $\bar{y}$. 
As for the methods described in \cref{sec-review_standardisation}, a major difference between previous implementations of divisive-normalisation is the extent of the neighbourhood, $\mathbb{P}$, defining the normalisation pool of each neuron.

Some methods calculate the weighted sum over both channels and space (C,H,W). However, this summation is restricted to a local spatial region \citep{Cirincione_etal22,Ren_etal17}.
\citet{Cirincione_etal22} used this form of divisive-normalisation as a front-end (part of an extended VOneBlock) to a ResNet applied to image classification with the ImageNetTiny data-set. They found this improved accuracy on the common corruptions data. \citet{Ren_etal17} found improved clean accuracy  on CIFAR10 and CIFAR100 when this form of divisive-normalisation was used in multiple layers of a ResNet.
Another method calculates the weighted sum over space only (H,W), but within a local neighbourhood. \citet{Hernandez-Camara_etal23} applied this form of divisive-normalisation in certain layers of a U-Net applied to image semantic segmentation, and found that this improved generalisation performance due to increased contrast and illumination tolerance.
Other implementations have calculated the weighted sum over channels only (C), but such that only a sub-set of channels contribute to the weighted sum. 
\citet{Krizhevsky_etal12} restricted the use of this form of divisive-normalisation to only the first two layers of an AlexNet. Others implemented modified versions of this method in more layers of AlexNet  and found improved clean test accuracy on ImageNet1k \citep{Pan_etal21,Miller_etal22}, and improved performance with noisy test data \citep{Miller_etal22}.

Our proposed method sums across all channels to calculate $\bar{y}$. Another major difference between our implementation and all others (except \citet{Miller_etal22}) is that we include the accelerating nonlinearity by setting $\alpha=2$. Previous work has mostly ignored this (\ie it has set $\alpha=1$), despite neuroscientific evidence supporting a value of  $\alpha$ of approximately two.






\subsection{Other Mechanisms of Competition}
\label{sec-review_other}
A number of alternative mechanisms have been used to produce competition in DNNs.
For example, a k-Winners-Take-All (k-WTA) activation function was found to improve adversarial robustness \citep{Xiao_etal19}. However, the improvement was modest and was caused by k-WTA destroying network gradient information. Hence, this method suffers from the limitations of other defences that rely on obfuscated gradients \citep{Athalye_etal18,Carlini_etal19}.
Setting to zero all but the k highest magnitude (positive and negative) values in each channel (\ie k-WTA applied across space separately in each channel), made a CNN learn more shape-biased representations \citep{Li_etal23}. 
Inhibition between neurons, in proportion the overlap between their weights, 
introduced into selected layers of ResNet50 produced small improvements in generalisation and larger improvements in robustness to weak adversarial attacks \citep{GaoSpratling25}.

\section{Methods}
\label{sec-methods}

\subsection{HAND: Homeostasis, Accelerating Nonlinearity, and Divisive-nomalisation}\label{sec-hand}

We implement our new activation function as follows:
\begin{equation}
  z=\frac{y}
  {\sigma + \bar{y}}  \;\;\;\;\;\;\;\;\;
  \bar{y}=w\sum_{i\in \mathbb{P}}y_i  \;\;\;\;\;\;\;\;\;
  y=\texttt{ReLU}(\texttt{BatchNorm}(x))^2
  \label{eq-hand}
\end{equation}
Where $\sigma$ and $w$ are learnt, scalar, non-negative parameters shared by all neurons in a feature-map, and the other symbols are as defined for \cref{eq-div-norm}.
Compared to the standard implementation of divisive-normalisation, \cref{eq-div-norm}, we make a number of changes which are described below.


Our aim was to implement competition between neurons that represent alternative interpretations of the same visual feature. We reasoned that this would require competition between all neurons at the same location in a feature map, and hence, set $\mathbb{P}$ to span across all channels. This is in contrast to previous versions (\cref{sec-review_div-norm}) that have limited the set $\mathbb{P}$ used for each neuron to a sub-set of neighbouring channels and/or a spatial neighbourhood. 
Preliminary experiments found that combining the activities of each neuron in the normalisation pool using separate, learnt, weights did not result in improved performance compared to using equal weighting. Hence, we reduced the number of learnt parameters, by summing the activities in the pool and multiplying this result by a single weighting factor, $w$. Furthermore, it was found that the same $w$ could be used for all output neurons without a reduction in performance, hence, a single weight could be learnt for the entire layer. Similarly, it was found that a single semi-saturation constant, $\sigma$, could be used for the entire layer. Hence, our implementation introduces only two learnt parameters per layer.
Modern CNNs can have a large number of channels. Hence, we initialised $w$ to a small value (equal to $\frac{1}{\text{num. channels}}$) to prevent initial values of $\bar{y}$ being large and suppressing all activity. The value of  $\sigma$ was initialised to one. Learning typically resulted in the value of $w$ increasing, so that the effects of normalisation became stronger over time. 


Increased separation between active and inactive neurons is produced by the accelerating non-linearity.
The exponent, $\alpha$, was fixed to a value of two which is the value typically estimated for this parameter from biological data. Preliminary experiments found that learning $\alpha$ could further improve clean and corrupt accuracy, but at the expense of reduced OOD rejection performance. 
Divisive-normalisation alone simply scales all activations equally rather than producing competition. This is because each of the neurons at the same location in a feature map are being normalised by the same value ($\sigma$ and $\bar{y}$ are the same for all such neurons).
However, divisive-normalisation does stabilize learning and encourages neurons to learn to respond to distinct patterns. This occurs because gradients that encourage a stronger output from a neuron will also encourage weaker outputs from other neurons at the same location in order to reduce $\bar{y}$.

One concern when introducing competition into a neural network is that it can increase the number dead neurons: ones that never respond as they always loose the competition. 
Even without competition mechanisms dead neurons are a concern in modern DNNs \citep{Pinto_etal25,Lu_etal20}. To address this concern we applied BatchNorm to the inputs to the activation function. This BatchNorm did not use the affine transformation. From a biological-perspective BatchNorm can be interpreted as performing a type of homeostasis: keeping each neuron within its stable operating conditions.


We replaced the standard activation function used in three CNN architectures with HAND. Some additional modifications were also made to avoid instabilities resulting from interactions between HAND and normalisation methods already implemented in some architectures.
For ConvNeXt \citep{Liu_etal22a} HAND replaced GELU in each block, and in the preceding convolutional layers the bias terms were removed, as was the affine transformation in the preceding LayerNorm.
For ResNet \citep{He_etal16} HAND replaced ReLU in the first two activation functions in each block and the BatchNorms preceding these activation functions were removed. 
For AlexNet \citep{Krizhevsky_etal12} HAND replaced the ReLUs following each convolutional layer, and no other modifications were made.

\subsection{Training Settings}
\label{sec-training_settings}
Performance was assessed for CNNs trained on ImageNet1k \citep{Russakovsky_etal15} and the long-tailed version of this data: ImageNetLT. For ImageNetLT the standard image sub-sets were used\footnote{\url{https://drive.google.com/drive/u/1/folders/1j7Nkfe6ZhzKFXePHdsseeeGI877Xu1yf}}, meaning that the number of samples per class ranged from 5 to 1280 (corresponding to an imbalance ratio of 256).
The training set-ups used were based on the standard pytorch recipes\footnote{\url{https://github.com/pytorch/vision/tree/main/references/classification}} for the architectures we used. Specifically, for training AlexNet  we used the SGD optimiser with a momentum of 0.9, weight decay of 1e-4, and a batch size of 128. The initial learning rate was 0.01 and this was reduced by a factor of 10 after $\frac{1}{3}$ and $\frac{2}{3}$ of the total training epochs.
For ResNet50 the set-up was the same except the batch size was 512 and the initial learning rate was 0.1 which was reduced $\frac{1}{4}$, $\frac{1}{2}$ and $\frac{3}{4}$ of the way through training.
For training ConvNeXt we used the AdamW optimiser \citep{LoshchilovHutter19} with weight decay of 0.05, except for the normalisation layers where no weight decay was used. The batch size was 512 and the peak learning rate was 0.001. The One-Cycle learning schedule \citep{SmithTopin18} was used with the warm-up period set to be 20 epochs or $\frac{3}{10}$ of the total training epochs, whichever was the smallest. Cross-entropy (CE) loss was used in all cases. Comparable training time information was obtained by training AlexNet on a single NVIDIA Tesla V100 GPU with 16GB of memory, while other networks were trained using four NVIDIA Tesla A100 GPUs with 40GB.

Two different training data augmentation schemes were used. The {\bf basic} augmentations consisted of RandomResizedCrop to 224 pixels and random horizontal flipping. The {\bf strong} augmentation scheme included the basic augmentations plus TrivialAug \citep{MullerHutter21}, RandErase with probability 0.1 \citep{Zhong_etal17}, mixup \citep{Zhang_etal18}, and cutmix \citep{Yun_etal19}. When using this scheme we also used label smoothing \citep{Szegedy_etal15b} in accordance with the pytorch standard training recipe.

To ensure the results generalised to other data-sets some additional experiments were performed with SVHN \citep{Netzer_etal11}, CIFAR10 \citep{Krizhevsky09}, CIFAR100 \citep{Krizhevsky09}, TinyImageNet (TinyIN), and long-tailed versions of CIFAR10 and CIFAR100. The long-tailed versions were created, as in previous work \citep{Cao_etal19,Liu_etal19,Wang21}, by removing training samples unequally from each class. Different degrees of imbalance were produced so that the ratio of the number of samples in the classes with the largest and smallest numbers was 100 (designated -LT100) or 10 (``-LT10''). These additional experiments were performed with ConvNeXt-tiny with the standard stem (which performs convolution with 4x4 kernels and a stride of 4) replaced by one more appropriate to small sized images (3x3 kernels and a stride of 1). This was  trained using the same training hyper-parameters as specified above, except the batch size was 128.

Some experiments with long-tailed data were performed with Logit-adjusted (LA) loss \citep{Menon_etal21}.
LA loss is a variation of CE loss where the logits are modified by a term that is proportional to the relative number of training samples in each class. Hence,
\begin{equation}
  \mathcal{L}_{LA} = - \log z_l 
  \label{eq-la_loss}
\end{equation}
where $l$ is the index corresponding to the correct class (\ie the ground-truth class label), and $\mathbf{z}$ is the result of applying the softmax function to the adjusted logits:
\begin{equation}
z_j=\frac{\exp({y_j+\log (p_j}))}{\sum_{i=1}^n\exp({y_i+\log (p_i)})}
\end{equation}
The term $p_j$ is the proportion of training samples in class $j$, i.e. $p_j={s_j}/{\sum_{i=1}^n s_i}$ where $s_j$ is the number of samples in class $j$, $\mathbf{y} \in \mathds{R}^n$ are the logits, and $n$ is the number of classes. A number of similar losses have been proposed which use alternative methods to adjust the logits \citep{Tan_etal20,Ren_etal20b,Cao_etal19}. However, LA loss has been found to produce better results than these alternatives and many other methods of dealing with class imbalance \citep{Menon_etal21,Zhao_etal24}.
In a few situations learning failed (there was either training collapse or no improvement in performance on the validation set) when using LA loss with the default training setup described above. Specifically, for the experiments reported in the third panel of \cref{tab-generalisation} training failed for ResNet50 with both ReLU and HAND, and for the standard versions of ConvNeXt-tiny and ConvNeXt-base. In these situations the initial learning rate was reduced by a factor of 0.1 which allowed training to complete successfully.


\subsection{Performance Metrics}\label{sec-metrics}
Performance was evaluated against a number of different criteria.
Firstly, the percentage of samples correctly classified from the ImageNet1k validation set (the  ``clean'' accuracy). Note, that this means that ImageNetLT-trained networks were tested with balanced data.
Secondly, the ability of trained networks to generalise to input distribution shifts was assessed by determining classification accuracy with the common corruptions data-set: ImageNet-C \citep{HendrycksDietterich19}. This data-set contains 18 different corruptions including different types of noise, blurring, synthetic weather conditions, and digital corruptions. As is typical in the literature, performance was evaluated by averaging performance over all the corruptions at all degrees of intensity.

A third performance metric was used to assess the ability of a network to distinguish known from unknown classes. This was evaluated using the 
Area Under the Receiver Operating Characteristic curve (AUROC) as this is a common choice in the literature \citep{Kirchheim_etal22,Chen_etal23,Cheng_etal23,Xu-Darme_etal23,Yang_etal22,Yang_etal23,Lee_etal22}. 
 AUROC is calculated separately for each unknown class data-set, evaluating how distinct the confidence scores produced by samples from the standard test-set are from the confidence scores produced in response to samples from the unknown class data-set. 
 The method for determining confidence was Maximum Logit Score (MLS) \citep{Vaze_etal22,Hendrycks_etal22b}. MLS defines the confidence that a sample is of a known class as the maximum response of the network output before any activation function is applied.
AUROC was calculated using the seven data-sets specified in \citep{Spratling_robustnessevaluation}, and the average AUROC across all seven data-sets was reported. These seven data-set containing unknown classes were Textures \citep{Cimpoi_etal14}, the iNaturalist 2021 validation set \citep{VanHorn_etal18}, the ImageNet-O data-set \citep{Hendrycks_etal19c}, 
and four data-sets containing synthetic images: (1) images containing random blobs, as used in \citep{Hendrycks_etal19}; (2) images in which each pixel intensity value was independently and randomly selected from a uniform distribution; (3) the images of the standard (clean) test set after a random permutation of all pixels; (4) the images of the clean test set after randomising the phase, in the Fourier domain, of each image.



Additional experiments to evaluate generalisation in ImageNet-trained models were performed with the  ModelvsHuman benchmark \citep{Geirhos_etal21}. This employs 17 data-sets, including multiple types of image corruption \citep{Geirhos_etal18}, edge-filtered images and silhouettes, Stylized-ImageNet and texture-shape cue conflict images \citep{Geirhos_etal18b}, and ImageNet-Sketch \citep{Wang_etal19}. As well as evaluating networks in terms of the accuracy in classifying these images, this benchmark also evaluates the consistency between the classification decisions made by a DNN and human subjects who have been tested using the same images (``error consistency'').  

The additional experiments performed using alternative training data-sets were evaluated in equivalent ways. Generalisation was assessed using MNIST-C \citep{MuGilmer19} (for SVHN-trained models), and CIFAR10-C,  CIFAR100-C, and TinyImageNet-C \citep{HendrycksDietterich19}  (for CIFAR10, CIFAR100, and TinyIN-trained models, respectively). For each training data-set, OOD Rejection was evaluated using the seven test-sets specified in \citet{Spratling_robustnessevaluation}.




Each experiment was repeated three times (using a different random weight initialisation and random presentation order of training samples), and the average performance across these three repeats is reported. To identify conditions where the difference in performance was statistically significant, performance produced by corresponding networks using HAND and the standard activation function were compared using the two-sample t-test (with $p<0.05$).

\section{Results}\label{sec-results}







\begin{figure*}[tbp]
  \begin{center}
\rotatebox{90}{\hspace*{12mm}\small Basic Augmentations}\hspace*{5mm}
\raisebox{42mm}{a}\includegraphics[scale=\figscale]{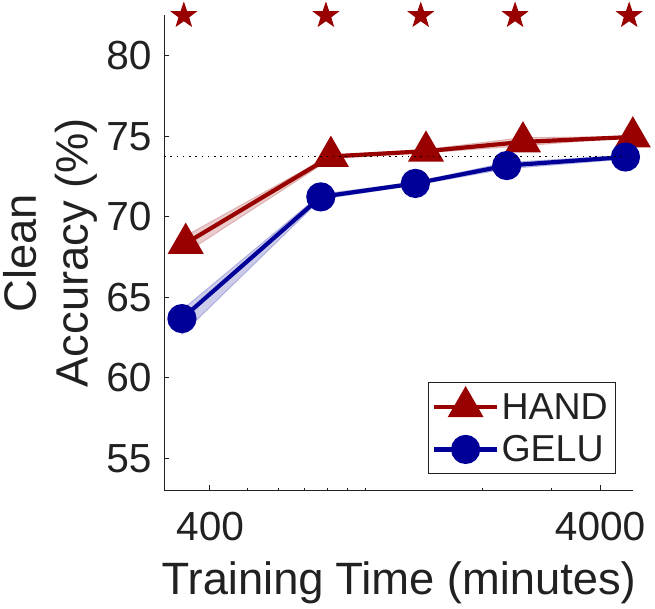}\hspace*{5mm}
\raisebox{42mm}{b}\includegraphics[scale=\figscale]{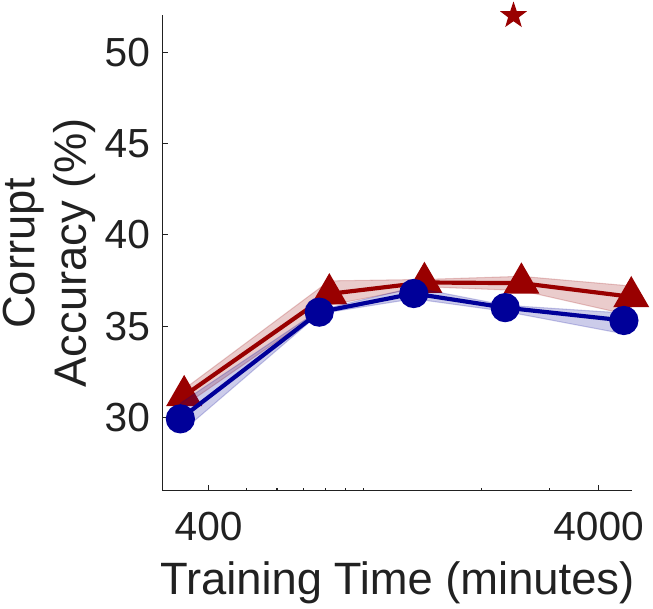}\hspace*{5mm}
\raisebox{42mm}{c}\includegraphics[scale=\figscale]{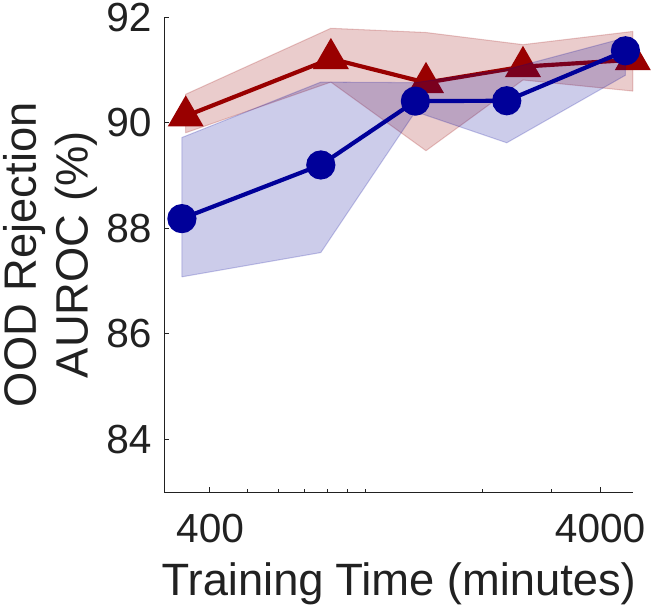}\\[2mm]
\rotatebox{90}{\hspace*{9mm}\small Strong Augmentations}\hspace*{5mm}
\raisebox{42mm}{d}\includegraphics[scale=\figscale]{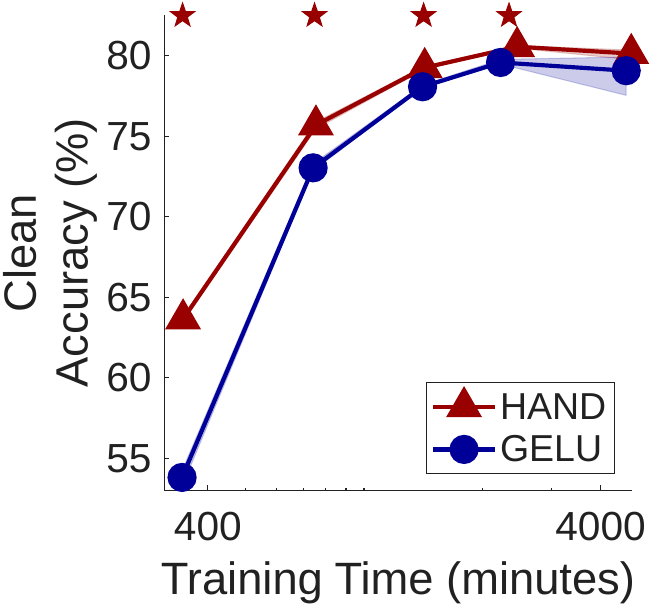}\hspace*{5mm}
\raisebox{42mm}{e}\includegraphics[scale=\figscale]{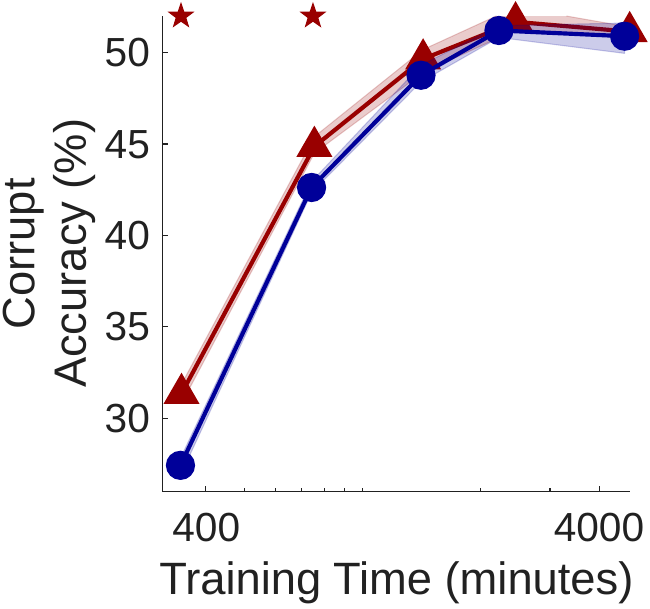}\hspace*{5mm}
\raisebox{42mm}{f}\includegraphics[scale=\figscale]{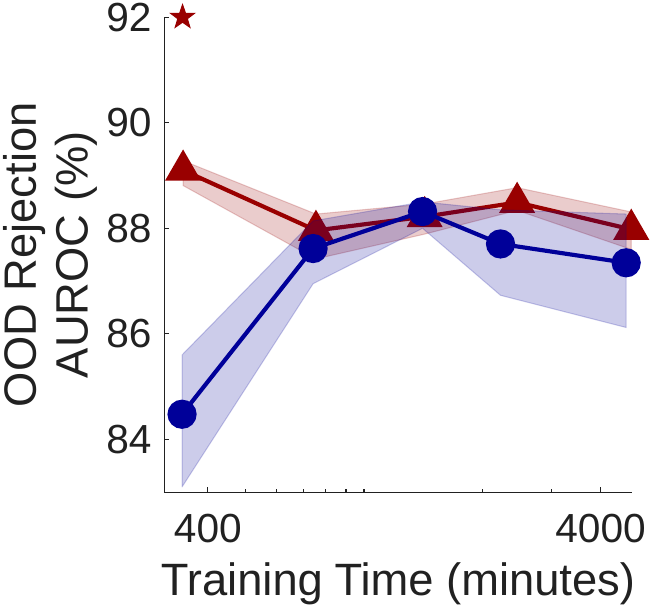}
\caption{Effects of training duration for ConvNeXt-tiny trained on ImageNet1k. The plotted points correspond to 10, 25, 50, 100, and 200 training epochs. Each shows performance of separate DNNs trained for that number of epochs, not the evaluations of checkpoints taken during the training of the same DNN. The solid lines show the mean values over three trials, and the shaded regions (which can be too narrow to be visible) indicate the minimum and maximum values recorded in any of the three trials. Stars indicate conditions where the difference in performance is statistically significant.}
\label{fig-resultsINepochs}
\end{center}
\end{figure*}

\subsection{Effects of Training Duration and Data Augmentation}
\label{sec-training_duration}

Using HAND as the activation function resulted in better performance particularly for short training times (\cref{fig-resultsINepochs}). On average a ConvNeXt-tiny incorporating HAND trained for 25 epochs on ImageNet1k with basic training data augmentation (\cref{fig-resultsINepochs}, top row) had clean accuracy of 73.76\%, corrupt accuracy of 36.77\%, and OOD rejection AUROC of 91.21\%. An unmodified ConvNeXt-tiny achieved 71.24\%, 35.77\%, and 89.20\% average performance on these metrics when trained in the same way. The unmodified network needed to be trained for 200 epochs to reach the same clean accuracy as the HAND network achieved after 25 epochs. Specifically, at 200 epochs the unmodified networks metrics were: 73.72\%, 35.31\%, and 91.36\%.
Replacing GELU with HAND in ConvNeXt-tiny increases the time taken for each epoch. Hence, for a fair comparison the results are plotted in terms of training time. The time required for the GELU-based network to reach the same clean accuracy as the HAND-based network achieved after 25 epochs was 5.7 times longer in terms of clock-time and 8 times longer in terms of training epochs.


At longer training times, the gap in performance narrows, as is expected if HAND acts as an inductive bias (see \cref{sec-introduction}). However, even at 200 training epochs the difference in clean accuracy is significant. Increasing the variety of training data, by using the strong augmentation scheme (\cref{sec-training_settings}), reduced the advantages produced by HAND and by 200 epochs had eliminated them (\cref{fig-resultsINepochs}, bottom row). As discussed in \cref{sec-introduction}, this is entirely consistent with the effects of inductive bias.
Comparing the corresponding graphs in the two rows of \cref{fig-resultsINepochs}, it can be seen that stronger training data augmentations improved generalisation performance, as measured by the clean and corrupt  accuracy, while reducing OOD rejection performance, and that this is the case for networks using both activation functions.  

\begin{figure*}[tbp]
  \begin{center}
\rotatebox{90}{\hspace*{20mm}\small CE Loss}\hspace*{5mm}
\raisebox{42mm}{a}\includegraphics[scale=\figscale]{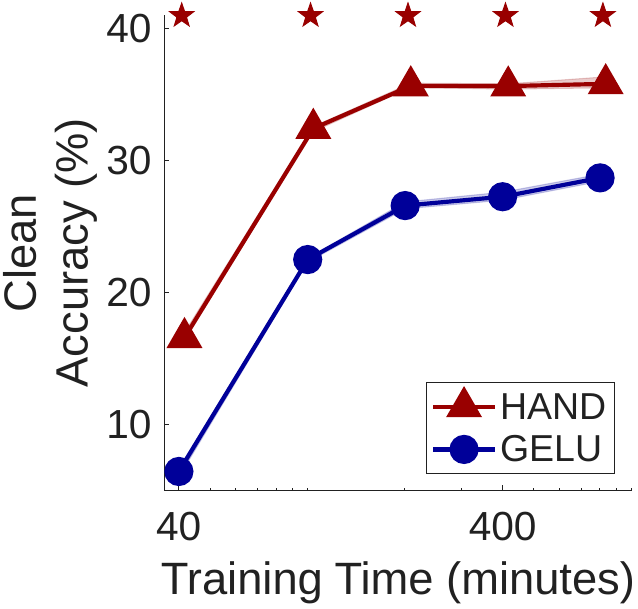}\hspace*{5mm}
\raisebox{42mm}{b}\includegraphics[scale=\figscale]{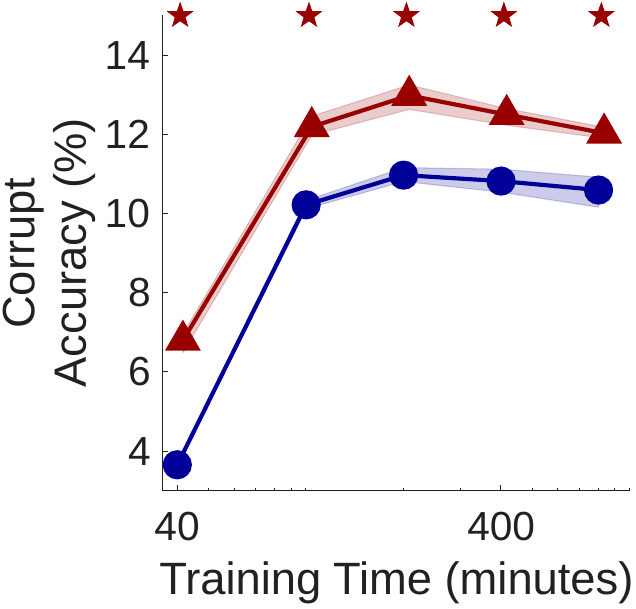}\hspace*{5mm}
\raisebox{42mm}{c}\includegraphics[scale=\figscale]{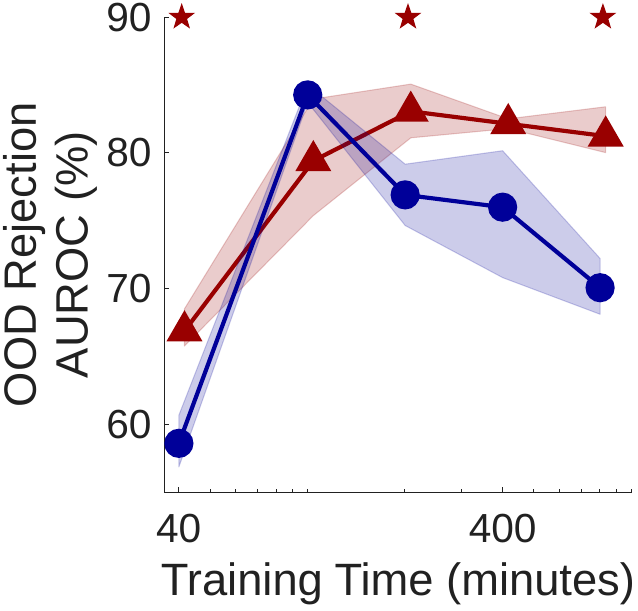}\\[2mm]
\rotatebox{90}{\hspace*{20mm}\small LA Loss}\hspace*{5mm}
\raisebox{42mm}{d}\includegraphics[scale=\figscale]{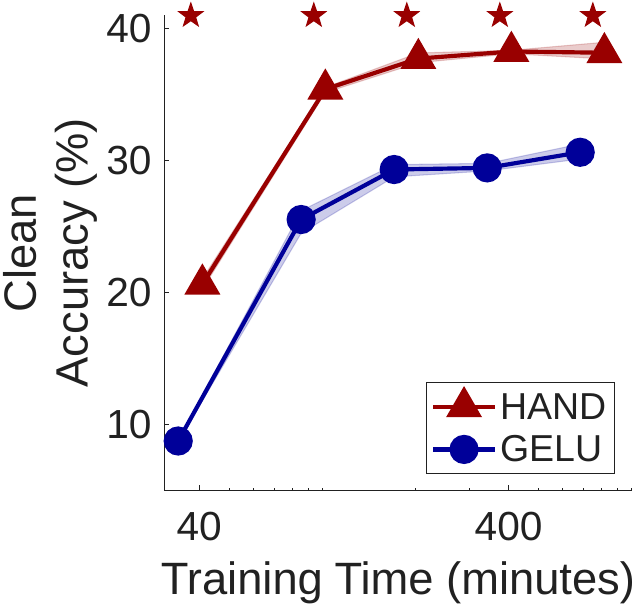}\hspace*{5mm}
\raisebox{42mm}{e}\includegraphics[scale=\figscale]{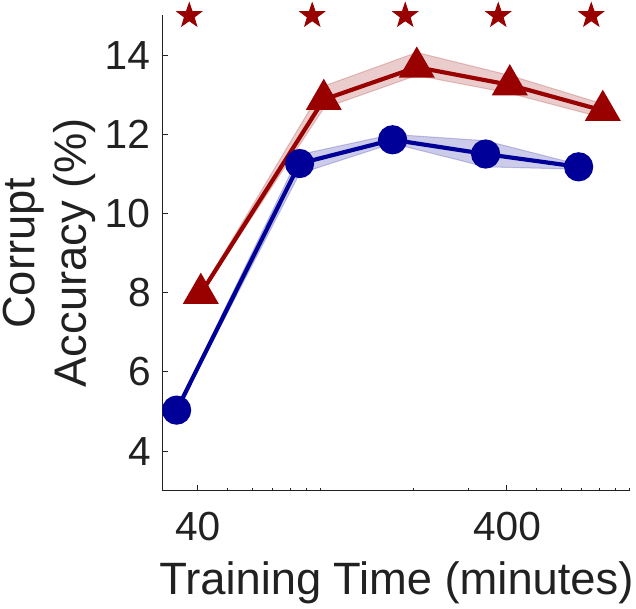}\hspace*{5mm}
\raisebox{42mm}{f}\includegraphics[scale=\figscale]{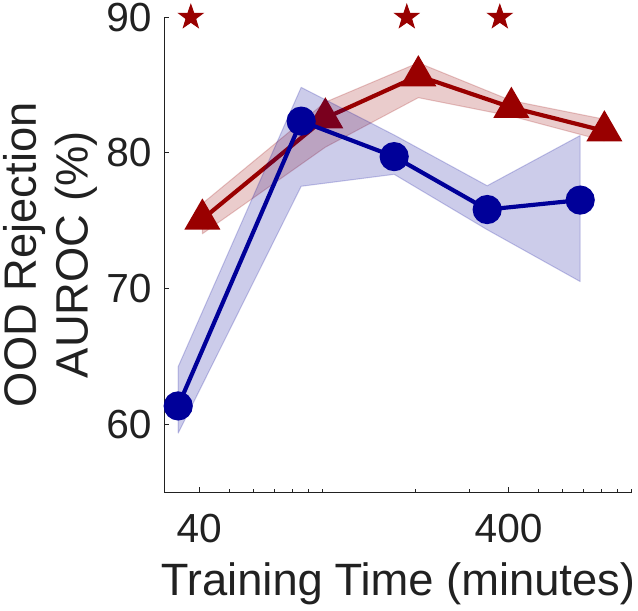}
\caption{Effects of training duration for ConvNeXt-tiny trained using basic augmentations on ImageNetLT. The plotted points correspond to 10, 25, 50, 100, and 200 training epochs. Each shows performance of separate DNNs trained for that number of epochs, not the evaluations of checkpoints taken during the training of the same DNN. The solid lines show the mean values over three trials, and the shaded regions (which can be too narrow to be visible) indicate the minimum and maximum values recorded in any of the three trials. Stars indicate conditions where the difference in performance is statistically significant.}
\label{fig-resultsLTepochs}
\end{center}
\end{figure*}

\subsection{Effects of Data Scarcity and Imbalance}
\label{sec-imbalance}



ImageNetLT is a version of ImageNet1k from which training samples have been removed, making training data more scarce. The number of samples removed from each class is unequal, making the training data imbalanced. When training with this data-set, the improvements produced by HAND were larger and were not reduced by a longer training duration (\cref{fig-resultsLTepochs}).
A ConvNeXt-tiny incorporating HAND trained for 25 epochs on ImageNetLT with basic training data augmentation had
clean  and corrupt accuracy unmatched by the unmodified network even after training for 200 epochs (\cref{fig-resultsLTepochs}a and b). The OOD rejection performance of the HAND-based network also exceed that of the GELU-based network after 50 training epochs (\cref{fig-resultsLTepochs}c).


Logit-adjusted (LA) loss \citep{Menon_etal21} is a loss function that is effective for training networks with imbalanced data. Repeating the previous experiment using LA instead of CE loss produced the results shown in the second row of \cref{fig-resultsLTepochs}. Comparing these results with those for CE loss in the first row of \cref{fig-resultsLTepochs} it can be seen that LA loss boosts clean and corrupt accuracy for the standard ConvNeXt-tiny. However, despite LA loss being a recent state-of-the-art method for dealing with long-tailed training data \citep{Zhao_etal24}, the improvements produced by LA are small compared to those obtained by using HAND. Furthermore, the clean and corrupt accuracy produced with HAND is also further boosted by using LA loss, meaning that the advantage over the standard, GELU-based, ConvNeXt-tiny is maintained. 

\subsection{Generalisation to Other CNN Architectures and Data-sets}
\label{sec-generalisation}

To ensure that the results presented above generalise from ConvNeXt-tiny, experiments were performed using two other CNN architectures and a larger version of ConvNeXt trained on ImageNet1k (\cref{tab-generalisation}, top panel) and ImageNetLT (\cref{tab-generalisation}, third panel).
It can be seen that the performance enhancement produced by HAND at short training durations with ImageNet1k and at longer training durations with ImageNetLT was consistent across CNN architectures and sizes. Furthermore, the more comprehensive evaluation provided by ModelvsHuman compared to common corruptions also showed a boost in generalisation performance from using HAND, and better alignment between the errors made by a HAND-based DNN and humans. The one exception was ResNet50 trained on ImageNetLT, where the results were mixed, suggesting HAND is less compatible with ResNet than ConvNeXt. 
We also found (results not shown) that HAND did not enhanced the performance of the vision transformer \citep{Dosovitskiy_etal20}. This was inline with our expectations, as visual transformers mix information from different parts of the whole image so that neurons do not have localised RFs in the same way as CNNs do.

\begin{table*}[tbp]
  \centering
  \caption{Comparison of HAND with standard activation functions across different CNN architectures and data-sets. Experiments performed with balanced training data-sets (top two panels) used 25 training epochs, CE loss, and basic data augmentations. Experiments performed using imbalanced training data-sets (lower two panels) used 100 training epochs, LA Loss, and strong data augmentations.
    Bold text indicates the best performance on each metric for each combination of training data-set and network architecture. Coloured text indicates that the difference in performance is statistically significant.}
    \label{tab-generalisation}
    \tabcolsep5pt
     \resizebox{\ifdim\width>\linewidth\linewidth\else\width\fi}{!}{%
      \begin{tabular}{lll *{3}{S[table-format=2.2(3)]}*{1}{S[table-format=1.3(3)]}*{1}{S[table-format=2.2(3)]}}\toprule
Training & Architecture &Activation & {Clean}         & {Corrupt}       & \multicolumn{2}{c}{{ModelvsHuman}} & {OOD rejection} \\
Data-set&              &Function   & {Accuracy (\%)} & {Accuracy (\%)} & {Accuracy (\%)} & {Error Consistency} & {AUROC (\%)} \\
        \midrule
\multirow{2}{*}{ImageNet1k}&\multirow{2}{*}{AlexNet}
&HAND         &\W51.05	(0.15)&\W20.78	(0.14)&\W41.67	(0.25)&\B0.165	(0.008)&\B 	84.72	(0.68)\\
&&ReLU        &  46.26	(0.23)&	 19.75	(0.26)&	 39.03	(0.57)&  0.159	(0.006)& 	78.46	(4.22)\\[\tabsecskip]
\multirow{2}{*}{ImageNet1k}&\multirow{2}{*}{ResNet50}                        
&HAND         &\B67.35	(0.48)&\W31.06	(0.44)&\B50.33	(0.32)&\B0.186	(0.005)& 	87.13	(2.27)\\
&&ReLU        &  65.51	(1.12)&	 29.20	(0.89)&	 49.33	(0.51)&  0.185	(0.001)&\B 	88.52	(0.44)\\[\tabsecskip]
\multirow{2}{*}{ImageNet1k}&\multirow{2}{*}{ConvNeXt-tiny}                   
&HAND          &\W73.76	(0.13)&\W36.77	(0.67)&\W55.00	(0.46)&\B0.192	(0.005)&\B   91.21	(0.52)\\
&&GELU         &  71.24	(0.16)&	 35.77	(0.10)&	 52.43	(0.35)&  0.182	(0.002)&     89.20	(1.62)\\[\tabsecskip]
\multirow{2}{*}{ImageNet1k}&\multirow{2}{*}{ConvNeXt-base}                   
&HAND          &\W74.94	(0.06)&\W40.21	(0.10)&\W56.73	(0.86)&\B0.190	(0.011)&\B	92.01	(0.22)\\
&&GELU         &  72.94	(0.19)&	 39.02	(0.12)&  53.83	(0.84)&  0.179	(0.003)&    90.27	(0.89)\\[\tabsecskip]
        \midrule
\multirow{2}{*}{SVHN}&\multirow{2}{*}{ConvNeXt-tiny}
&HAND         &\W95.75	(0.06)&	  57.24	(1.78)&n/a&n/a&	  94.51	(0.79) \\
&&GELU        &  94.47	(0.02)&	\B59.20	(2.30)&n/a&n/a&	\F96.55	(0.74)\\[\tabsecskip]
\multirow{2}{*}{CIFAR10}   &\multirow{2}{*}{ConvNeXt-tiny}       
&HAND         &\W91.38	(0.23)&	\B75.86	(0.60)&n/a&n/a&	\B91.30	(1.01)\\
&&GELU        &  88.49	(0.43)&	  75.53	(0.27)&n/a&n/a&	  89.82	(0.82) \\[\tabsecskip]
\multirow{2}{*}{CIFAR100}     &\multirow{2}{*}{ConvNeXt-tiny}    
&HAND         &\W69.31	(0.98)&	\W49.62	(0.25)&n/a&n/a&	\B81.42	(1.36)\\
&&GELU        &  62.14	(0.69)&	  46.09	(0.21)&n/a&n/a&	  79.42	(1.72)\\[\tabsecskip]
\multirow{2}{*}{TinyIN} &\multirow{2}{*}{ConvNeXt-tiny}    
&HAND         &\W57.72	(0.17)&\W25.29	(0.09)&n/a&n/a&	  \W73.80	(1.37)\\
&&GELU        &  50.62	(0.84)&	 19.25	(0.37)&n/a&n/a&	  	 63.54	(4.14)\\[\tabsecskip]
        \midrule
\multirow{2}{*}{ImageNetLT}&\multirow{2}{*}{AlexNet}
&HAND          &\W27.69	(0.66)&\W12.00	(0.09)&\W34.70	(0.66)&\W0.128	(0.002)&\W77.31	(1.41)\\
&&ReLU         &  16.59	(0.55)&	 8.30	(0.26)&  24.40	(0.95)&  0.083	(0.006)&  60.89	(0.74)\\[\tabsecskip]
\multirow{2}{*}{ImageNetLT}&\multirow{2}{*}{ResNet50}                        
&HAND          &  31.90	(0.54)&\B14.02	(0.18)&  36.23	(0.57)&    0.129	(0.004)&			68.50	(7.38)\\
&&ReLU         &\B33.23	(1.15)&  13.78	(0.56)&\F37.70	(0.36)&\B  0.130	(0.008)&\B			82.20	(0.52)\\[\tabsecskip]
\multirow{2}{*}{ImageNetLT}&\multirow{2}{*}{ConvNeXt-tiny}                   
&HAND          &\W50.48	(0.57)&\W23.04	(0.37)&\W47.93	(1.07)&\W0.168	(0.003)&\W84.45	(0.48)\\
&&GELU         &  23.15	(0.87)&	 10.32	(0.33)&	 29.80	(1.51)&  0.101	(0.006)&  75.80	(1.99)\\[\tabsecskip]
\multirow{2}{*}{ImageNetLT}&\multirow{2}{*}{ConvNeXt-base}                   
&HAND          &\W51.49	(0.38)&\W24.62	(0.73)&\W48.67	(0.59)&\W0.166	(0.007)&\W85.67	(1.11)\\
&&GELU         &  31.52	(0.35)&  14.45	(0.09)&  37.10	(0.20)&  0.125	(0.006)&  77.81	(1.38)\\[\tabsecskip]
\midrule
CIFAR10&\multirow{2}{*}{ConvNeXt-tiny}       
&HAND          &\W78.64	(0.17)&	\W66.62	(0.41)&	n/a&n/a&  \W73.55	(4.25)\\
-LT100&&GELU   &  68.33	(0.52)&	  61.02	(0.64)&	n/a&n/a&    63.57	(2.05)\\[\tabsecskip]
CIFAR10   &\multirow{2}{*}{ConvNeXt-tiny}                            
&HAND          &\W90.66	(0.05)&	\W79.99	(0.24)&	n/a&n/a&  \B80.21	(5.54)\\
-LT10&&GELU    &  88.06	(0.33)&	  78.87	(0.48)&	n/a&n/a&    74.81	(0.95)\\[\tabsecskip]
CIFAR100   &\multirow{2}{*}{ConvNeXt-tiny}                       
&HAND          &\W47.54	(0.90)&	\W37.27	(0.98)&	n/a&n/a&  \W75.73	(5.46)\\
-LT100&&GELU   &  34.90	(1.00)&	  29.06	(0.97)&	n/a&n/a&    50.95	(3.07)\\[\tabsecskip]
CIFAR100     &\multirow{2}{*}{ConvNeXt-tiny}                     
&HAND          &\W64.68	(0.59)&	\W51.51	(0.59)&	n/a&n/a&  \W81.98	(2.25)\\
-LT10&&GELU    &  57.45	(0.75)&	  46.95	(0.72)&	n/a&n/a&    72.35	(1.05)\\[\tabsecskip]
\bottomrule
       \end{tabular}}\end{table*}


To ensure that the results presented in the preceding sections also generalise from ImageNet1k to other data-sets, experiments were performed with other balanced training data-sets (\cref{tab-generalisation}, second panel) and imbalanced training data (\cref{tab-generalisation}, bottom panel). In all cases, substituting the standard activation function for HAND produced similar or improved performance on all metrics. For SVHN the effects of HAND were mixed. However, this is an easy to learn data-set and performance has plateaued long before 25 training epochs. Hence, this result shows the effects of HAND after plenty of training epochs, where the effects of an inductive bias are expected to be weak. For the harder data-sets (CIFAR10, CIFAR100, and TinyIN) HAND improves performance on all metrics. HAND also improved performance on all metrics for the additional long-tailed data-sets, and this improvement was almost always statistically-significant.

\begin{table*}[tbp]
  \centering
    \tabcolsep20pt
  \caption{Results of the ablation study. Training performed with ImageNet1k for 25 epochs with basic data augmentation. Results are also included for a version of AlexNet in which BatchNorm was applied prior to ReLU, to provide a fairer baseline with which to compare our results. Bold text indicates the best performance on each metric for each network architecture. }
    \label{tab-ablation}
     \resizebox{\ifdim\width>\linewidth\linewidth\else\width\fi}{!}{%
      \begin{tabular}{ll *{3}{S[table-format=2.2(2)]}}\toprule
Architecture &Activation & {Clean}         & {Corrupt}       & {OOD rejection} \\
             &Function   & {Accuracy (\%)} & {Accuracy (\%)} & {AUROC (\%)} \\
        \midrule
\multirow{6}{*}{AlexNet}
&HAND                &\B51.05	(0.15)&\B20.78	(0.14)&\B84.72	(0.68)\\
&HAND w/o BatchNorm  &  \multicolumn{2}{c}{learning failed}\\
&HAND w/o squaring   &  48.41	(0.51)&	18.51	(0.05)&	81.64	(1.42)\\
&HAND w/o div-norm   &  \multicolumn{2}{c}{learning failed}\\
&BatchNorm + ReLU    &  49.76	(0.03)&	19.54	(0.12)&	81.94	(2.82)\\
&ReLU                &  46.26	(0.23)&	19.75	(0.26)&	78.46	(4.22)\\[2\tabsecskip]
\multirow{5}{*}{ConvNeXt-tiny}
&HAND                &  73.76	(0.13)&	 36.77	(0.67)&  91.21	(0.52)\\
&HAND w/o BatchNorm  &  71.87	(0.09)&	 36.27	(0.29)&	 89.84	(0.82)\\
&HAND w/o squaring   &  71.77	(0.14)&	 32.89	(0.38)&	 90.97	(0.71)\\
&HAND w/o div-norm   &\B74.11	(0.19)&\B38.20	(0.26)&\B91.24	(0.90)\\
&GELU                &  71.24	(0.16)&	 35.77	(0.10)&	 89.20	(1.62)\\
\bottomrule
      \end{tabular}
      }
\end{table*}

\subsection{Ablation Study}
\label{sec-ablation}

HAND consists of three components: Homeostasis (implemented using BatchNorm), an Accelerating Nonlinearity (implemented by squaring the output of ReLU), and Divisive-normalisation. To evaluate the contribution of each of these components, two architectures were trained with ablated versions of HAND that were each missing one component (see \cref{tab-ablation}).
BatchNorm was found to be essential for AlexNet, and produced improvements in the clean accuracy and OOD rejection performance with ConvNeXt-tiny. Removing the squaring reduces performance for both architectures on all metrics. The divisive-normalisation component was essential for stable learning with AlexNet, but reduced overall performance for ConvNeXt-tiny: however, this reduction was not statistically-significant for any of the metrics used. 
These results suggest that the accelerating non-linearity is the component that contributes most strongly to the improved performance produced by HAND, but that for some architectures the other components are essential to stabilise learning.
Divisive normalisation has previously been found to perform an important role in maintaining the stability of neural activity during inference in recurrent networks \citep{Rawat_etal24,Morone_etal26}. Here, we find that it plays a role in training stability in feed-forward networks as well.

\subsection{Comparison with Previous Divisive-Normalisation Implementations}
\label{sec-compare_div-norm}

Existing normalisation layers (\cref{sec-review_standardisation}), and almost all 
previous implementations of divisive-normalisation applied to DNNs (\cref{sec-review_div-norm}), do not include an accelerating nonlinearity. Our ablation results show that when this component is missing the improvements in performance over unmodified networks are small, which is consistent with the results produced by previous methods that have been inspired by work on divisive-normalisation.

The previous implementations of divisive-normalisation that have been tested with ImageNet1k have been incorporated into AlexNet. These methods have not been shown to scale-up to modern architectures, nor have they been rigorously evaluated in terms of generalisation and OOD rejection. \citet{Miller_etal22}, the only previous method to use an accelerating nonlinearity, reports a clean accuracy after 90 training epochs of 59.61\%. However, this was produced using an AlexNet with 96, 256, 384, 384, and 256 channels in its convolutional layers. Our results in earlier sections were produced using the standard pytorch implementation of AlexNet that contains 64, 192, 384, 256, and 256 channels. Using HAND with the larger version of AlexNet and the same training setup as used in \citet{Miller_etal22} we obtain almost identical clean accuracy: 59.60\%. \citet{Pan_etal21} claim higher clean accuracy ($\approx$64\%). However, they not only used the version of AlexNet with more convolutional filters, but also added two additional fully-connected layers. We did not test HAND with this bespoke architecture, but note that our ConvNeXt-tiny models with HAND can produce far better clean accuracy (73.76\%) than reported by \citet{Pan_etal21} using an architecture containing far fewer parameters. 



\section{Discussion}
\label{sec-discussion}

The bias-variance trade-off \citep{Geman_etal92} tells us that machine learning systems containing more inductive bias should require fewer training samples and should generalise more successfully. However, this principal does not tell us what are the appropriate inductive biases for different tasks. Here, we show that competition between neurons representing the same image patch, implemented using divisive-normalisation, is an appropriate inductive bias for image classification tasks.  Our results show that this method, implemented as a new activation function (HAND), produces large improvements in performance, especially when using few training epochs, or limited training-time augmentations, or few training samples for certain classes.

The success of our implementation of divisive normalization relies on the inclusion of the accelerating non-linearity and the homeostasis mechanism preceding the normalization. We observe that the squaring operation and the BatchNorm are both more important for performance than the divisive normalization, which also explains the relatively weak performance of previous implementations. Both accelerating non-linearity \citep{Albrecht_etal02,AlbrechtGeisler91,Heeger92b} and homeostasis mechanisms that keep neurons within their dynamic range \citep{Davis13,TurrigianoNelson00,DavisBezprozvanny01} are well supported by neuroscientific data, too. Thus, the improvements we include are still strongly inspired by biological neural networks and may further improve similarity between DNNs and biological neural networks.

Increasing inductive bias may have an important role to play in addressing several issues with deep learning, such as the exponentially increasing computational requirements \citep{Sevilla_etal22} which undermine AI sovereignty and democracy and lead to unsustainable increases in environmental impact \citep{Strubell_etal20,Thompson_etal20}.
In addition, the need for ever more training data reduces the ability to curate it: increasing the risk that the data contains, and hence the model learns, prejudices. Furthermore, this approach to improving AI seems to be reaching its limits, if claims that AI is running out of training data are to be believed \citep{Jones24}.

A limitation of increasing inductive bias is that the resulting system is less general: it incorporates constraints that are specific to certain tasks and inappropriate for others. However, the fact that a single image location can only contain one image feature, and hence have one interpretation, seems to be a prior relevant to all vision tasks. Therefore, it seems likely that HAND could also improve performance in other vision-based applications. Verifying this is left to future work. We have successfully integrated HAND into existing CNNs. However, better performance might be possible with architectures optimised especially to fully exploit competition. Further work might also consider improved methods of implementing competition. For example, stronger methods of competition, that are more effective at suppressing responses that are poorer matches to the input, might further improve the ability to reject samples from unknown classes.



\section*{Acknowledgements}
The simulations were performed using the Luxembourg national supercomputer MeluXina and the HPC facilities of the University of Luxembourg \citep{Varrette_etal22}.
The authors gratefully acknowledge the LuxProvide and University of Luxembourg teams for their expert support.

\printbibliography[]





\end{document}